# An Improved Deep Learning Model for Word Embeddings Based Clustering for Large Text Datasets

Vijay Kumar Sutrakar[#$] and Nikhil Mogre[#@]

*Aeronautical Development Establishment, Defence Research and Development Organisation, New Thipasandara Post, Bangalore, Karnataka, India 560093*
[@]nikhilmogre1998@gmail.com, [$]vks.ade@gov.in
[#] all authors contributed equally

**Abstract.** In this paper, an improved clustering technique for large textual datasets by leveraging fine-tuned word embeddings is presented. WEClustering technique is used as the base model. WEClustering model is further improvements incorporating fine-tuning contextual embeddings, advanced dimensionality reduction methods, and optimization of clustering algorithms. Experimental results on benchmark datasets demonstrate significant improvements in clustering metrics such as silhouette score, purity, and adjusted rand index (ARI). An increase of 45% and 67% of median silhouette score is reported for the proposed WEClustering_K++ (based on K-means) and WEClustering_A++ (based on Agglomerative models), respectively. The proposed technique will help to bridge the gap between semantic understanding and statistical robustness for large-scale text-mining tasks.

## 1 Introduction

Recently, the explosion for textual data in digital form posed significant challenges for efficient information retrieval and processing. For example, Elsevier's repository contains over 37,000 articles on COVID-19 alone, and the number of articles published in English-language journals continues to grow [1, 2]. A basic data grouping approach, clustering is frequently utilized in applications including document organization, social news clustering, and web search result clustering. It is also used as an initial step to tasks like sentiment analysis, topic extraction, and multi-document summarization. Bag of words (BOW) model [3] that utilizes scoring schemes like term frequency (TF) or term frequency inverse document frequency (TF-IDF) [4] to numerically represent documents is a basis of traditional document approaches for clustering. However, such approaches have major limitations. It fails to capture semantic relationships, struggle with polysemy and synonymy, and face the curse of dimensionality, particularly for large datasets with high sparsity [5, 6, 7, 8]. Ontology-based solutions like WordNet [9,10] partially address semantic issues; however, it is limited by language coverage and design constraints. To tackle these challenges, word embeddings such as Word2Vec [11], GloVe [12], and FastText [13] provide dense, distributed representations of words. However, these embeddings are static and do not account for context-specific meanings. Recently, Bidirectional Encoder Representations from Transformers (BERT) [14] has emerged as a breakthrough model, generating contextual embeddings that adapt based on input context. Fine-tuning BERT on domain-specific datasets further enhances its ability to capture nuanced semantics, making it a robust tool for clustering and other text mining tasks.

Recently, WEClustering [15] technique is proposed for document clustering that integrates word embeddings [16, 17, 18, 19] and clustering methods to improve clustering performance. It begins with pre-processing, where documents are converted to lowercase and split into sentences, followed by tokenization, stop-word removal, and punctuation filtering. BERT is used to extract word embeddings, creating 1024-dimensional vectors for each word. These embeddings are clustered using Mini-Batch K-Means to form word clusters. Next, a concept-document (CD) matrix is generated by mapping TFIDF scores of words to their respective cluster centres. The resulting CD matrix is normalized and clustered using algorithms like Agglomerative Clustering or K-Means to produce document clusters [20].

In this paper, an improved document clustering technique i.e. WEClustering++ that leverages word embeddings derived from a fine-tuned BERT model is proposed. The proposed model addresses the issue of dimensionality, incorporates contextual semantics, and demonstrates high accuracy. The effectiveness of the proposed approach is demonstrated using multiple datasets and performance metrics, highlighting its suitability for large-scale text datasets. The details of proposed model are provided in Section 2. Section 3 discussed results and discussions followed by concluding remarks in Section 4.

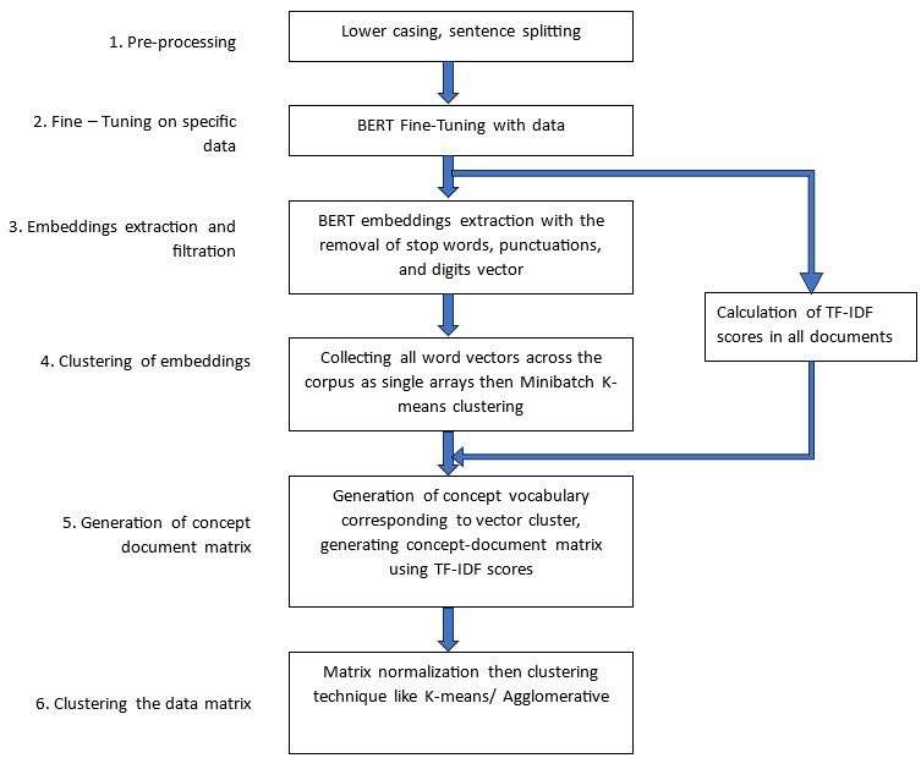

**Fig.1**. Flow chart of improved deep learning model for word embedding based clustering for large text datasets.

## 2 Methodology

### 2.1 Model details

The proposed model builds upon the WEClustering framework [15] by incorporating fine-tuning of the BERT model, enabling the system to specialize for specific domains or applications. This enhancement allows the model to capture domain-specific nuances in text, making the clustering process more precise and relevant. However, the foundational steps of WEClustering [15] are retained. The details of finetune of embeddings are shown in Fig. 1.

The first phase, preprocessing, involves preparing the text data for processing by the fine-tuned BERT model. This begins with lowercasing all text to ensure uniformity, followed by sentence splitting, as BERT operates at the sentence level to preserve contextual relationships. In the second phase, the pre-processed text is then ready for the fine-tuned BERT model, which generates embeddings that are contextually rich and domain-specific, aligning with the semantics of the targeted application. The third phase, embedding extraction and filtration, involves passing the pre-processed text through the finetuned BERT model. Unlike the original WEClustering [15], which uses a general pre-trained BERT model, the proposed model employs a domain-specific fine-tuned version. This enables better capture of word meanings and relationships relevant to the domain. The filtration step removes noise by excluding embeddings associated with stop words, punctuation, and digits, resulting in a refined set of embeddings that retain both statistical and semantic relevance.

The fourth phase focuses on clustering word embeddings. Mini-Batch K-Means [21] is used to group embeddings into clusters, each representing a concept or theme derived from the text. Elbow Method [22] is used for obtaining optimal cluster number, ensuring that the generated concepts are meaningful and representative of the corpus. By leveraging fine-tuned embeddings, these concepts become more cohesive and domain-relevant compared to those in the original WEClustering framework. This step also reduces the vocabulary size, transforming it from tens of thousands of words to a concise set of concepts. In the fifth phase, a Concept-Document (CD) matrix is constructed to represent documents in terms of the identified concepts rather than individual words. A scoring mechanism combining TF-IDF with the frequency of words from each concept calculates the relevance of concepts to each document. The fine-tuned embeddings ensure that these concepts are semantically aligned with the documents, creating an accurate and meaningful matrix representation.

In the final step, the documents are clustered using algorithms like K-Means [20] or Agglomerative Clustering [23], based on the CD matrix. The dimensionality of the input is substantially reduced compared to traditional word-based approaches, leading to more cohesive and well-separated clusters. Fine-tuned embeddings enhance the cohesion and representational accuracy of the clusters, ultimately resulting in well-defined and meaningful document groupings. The improved models developed in the present work are called as WEClustering_K++ (based on K-means) and WEClustering_A++ (based on Agglomerative clustering) in rest of the document. This comprehensive framework demonstrates the advantages of integrating domain-specific fine-tuning into the existing WEClustering process [15].

## 2.2 Model Parameters

In general, two BERT models are available for generating embedding vectors in the embedding extraction phase. The BERT-small model [24] generates embeddings of size 768 and is available in both case-sensitive and case-insensitive versions. The BERT-large model generates embeddings of size 1024 and is available only as a case-sensitive version. In the present work, the BERT-large model, fine-tuned on domain-specific datasets, is utilized. The fine-tuning ensures that the generated embeddings are not only rich in contextual meaning but also more aligned with domain-specific semantics, enhancing the clustering process. The use of higher-dimensional embeddings further improves the representation of word semantics.

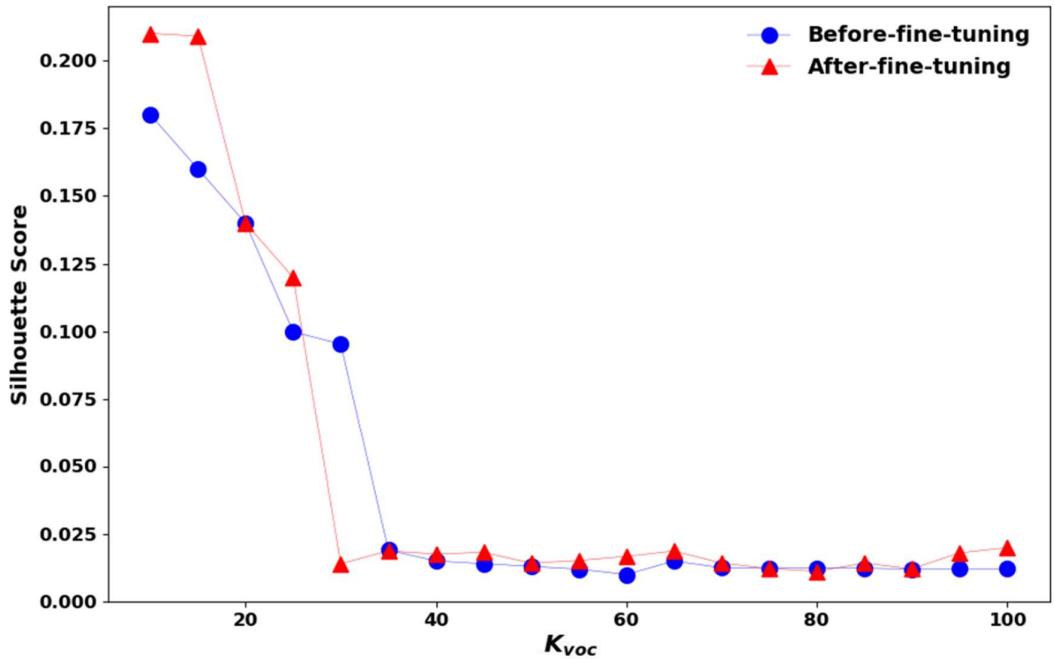

**Fig.2.** An example of Articles-253 dataset for finding $K_{voc}$ using Elbow method [22]

In the word embeddings phase, the Mini-Batch K-Means algorithm [21] is employed for clustering. The two key parameters in this phase are: (i) the number of word clusters ($K_{voc}$) and (ii) the batch size ($b$). The value of $K_{voc}$ is determined using the Elbow method [22], which ensures optimal clustering. For example, for the Articles-253 dataset, the value of $K_{voc}$ is determined as 35 using [25] and the present work the $K_{voc}$ for fine tuning is obtained 30, as shown in Fig. 2. The remaining model's $K_{voc}$ values are shown in Fig. 3. The incorporation of fine-tuned embeddings often leads to smaller, more cohesive clusters, reflecting the improved semantic alignment of the words. The batch size parameter, $b$ is chosen to balance execution efficiency and accuracy (taken from [25]).

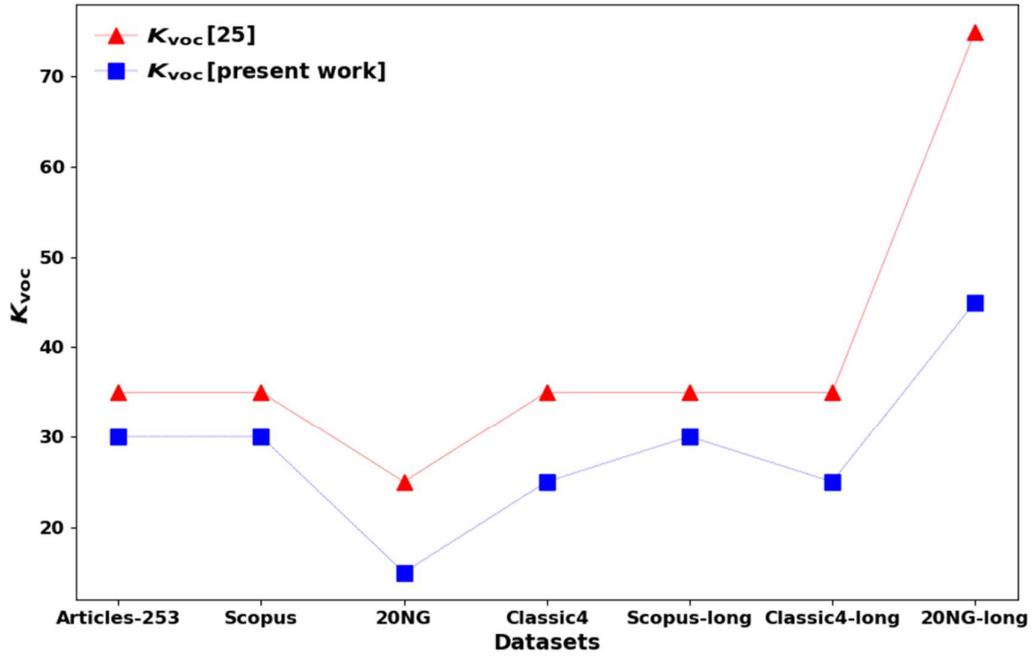

**Fig.3.** $K_{voc}$ considered for different datasets

In the final phase, document clustering is performed using algorithms such as K-means or agglomerative clustering. For agglomerative clustering, ward linkage is used as the linkage criterion. In the K-means clustering, the number of clusters ($c$) corresponds to the number of categories in the dataset. To improve the initialization of centroids, the K-means++ method is utilized [26], and the algorithm is executed 20 times to ensure the best clustering results.

The integration of fine-tuning in WEClustering significantly impacts the value of $K_{voc}$. Finetuned embeddings provide better semantic alignment and reduce redundancy, resulting in more meaningful and cohesive word clusters. Consequently, $K_{voc}$ decrease compared to the original WEClustering approach, as the finer embeddings better represent the underlying concepts in the text.

### 2.3 Datasets

The evaluation of an improved deep learning model for word embedding based clustering for large text dataset is conducted using seven benchmark datasets spanning various domains and sizes (refer [27] and [28] and Fig. 4 for further details). These datasets include collections of research articles, news articles, and other domain-specific texts. The details of each dataset are given next.

1. Articles-253: A collection of 253 research articles across five categories, such as "Mobile Computing," "Political Science," and "Weather Review," including titles, abstracts, and references.
2. Scopus: Contains 500 articles, evenly distributed into five categories like "Concrete," "Hyperactivity," and "Tectonic Plates," with titles and abstracts sourced from the Scopus database.
3. 20NG: A subset of the 20 Newsgroups dataset with 700 documents from four categories: "Atheism," "Religion," "Graphics," and "Space."
4. Classic4: Comprises 800 research articles across three main domains—"Aerodynamics," "Medical," and "Algorithms."
5. Scopus-long: A larger version of the Scopus dataset with 2800 articles spread across seven categories, including "Neural Networks," "Protons," and "Photosynthesis."
6. Classic4-long: An expanded version of Classic4 with 3891 documents.
7. 20NG-long: A larger subset of 20 Newsgroups with 8131 documents spanning nine categories like

8. "Motorcycles," "Politics," "Hockey," and "Electronics."

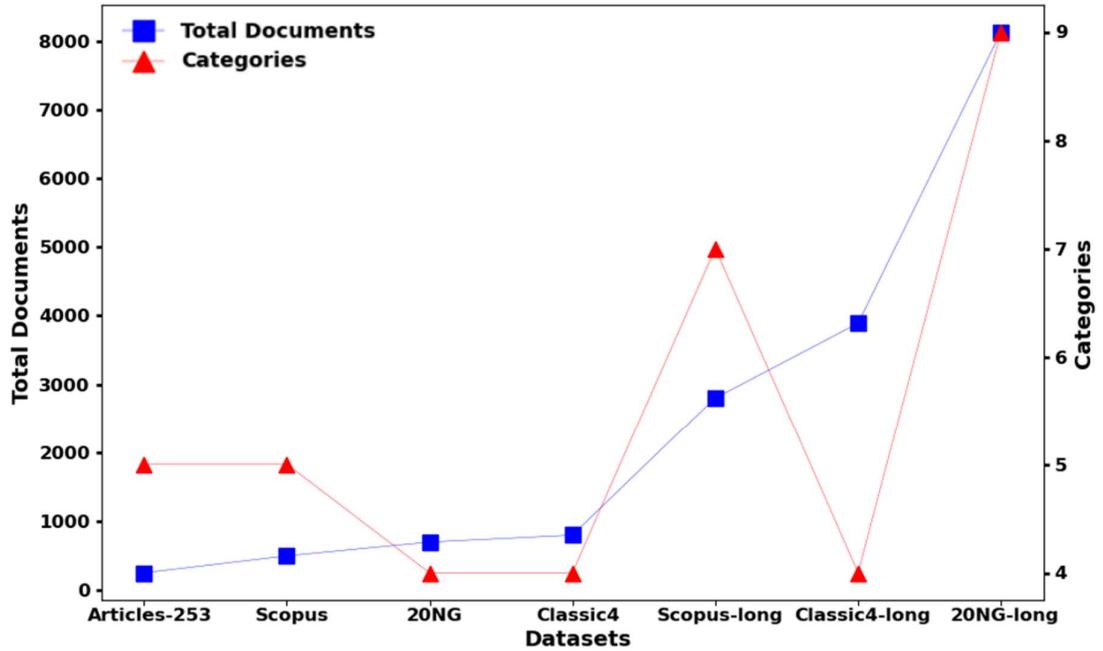

**Fig.4.** Total documents and categories considered for different data sets (taken from [27] and [28])

### 2.4 Performance Metrics

Clustering quality can be evaluated using various metrics that are broadly classified into external and internal categories, depending on the availability of ground truth labels. External metrics are used when true labels are available, while internal metrics are applicable when true labels are not known. In the present work, three metrics are used to assess the quality of clustering, i.e.(a) silhouette coefficient, (b) adjusted rand index (ARI), and (c) purity.

The silhouette coefficient is an internal metric that evaluates how well-separated and compact the clusters are. It is particularly useful when true labels are unavailable. Its values range from -1 to +1, where higher values indicate that clusters are dense and well-separated, while values closer to -1 suggest poor clustering [29]. The ARI is an external metric used to assess clustering quality when true labels are present. ARI measures the similarity between the clustering results and the true labels while adjusting for random chance. A score near +1 indicates strong agreement with the true labels, whereas a score near -1 signifies that the clustering is random or completely dissimilar to the actual labels [30]. Purity, another external metric, measures clustering quality by determining how well each cluster corresponds to a single class. Higher purity values indicate that the clusters align well with the true class labels [31]. Together, these metrics provide a comprehensive evaluation of clustering performance, enabling the assessment of both internal structure and alignment with external benchmarks when true labels are available.

## 3 Results and discussions

The WEClustering_K++ method exhibits notable enhancements in silhouette coefficient over WEClustering_K[15] across all datasets, as shown in Fig. 5(a) and Table 1. In the Articles-253 dataset, the proposed model achieves a silhouette score of 0.65, a significant 44.4% increase over WEClustering_K[15]. Classic4 also benefits, with the score rising from 0.21 to 0.46, reflecting a 119% improvement. Classic4-long experiences a boost from 0.238 to 0.385, equating to a 61.7% increase. In the Scopus dataset, the improvement stands at 41.5%, while Scopus-long sees a 36.8% rise. The most striking gain is in the 20NG-long dataset, where the silhouette score surges from 0.043 to 0.25, an impressive 481% increase. Even in the 20NG dataset, a smaller but noticeable improvement of 2% is recorded. Similarly, WEClustering_A++ surpasses WEClustering_A[15], showing a 53.5% gain (0.43 to 0.56) in Articles-253, an 88% increase in Classic4, and an 86% rise in Classic4-long. Scopus and Scopus-long also benefit, with improvements of 67.5% and 32%, respectively. The 20NG dataset improves by 42.8%, underscoring the superior performance of the proposed approach. Also, an increase of median silhouette score metrics of 45% and 67% is obtained for proposed WEClsutering_K++ and WEClustering_A++ compared to the state of art model, i.e. WEClustering_K[15] and WEClustering_A[15], respectively. In the case of WEClustering_K++, minimum increase of silhouette score metrics is +2% for 20NG dataset, while the maximum increase of silhouette score metrics is +481% for 20NG-long dataset. In the case of WEClustering_A++, minimum increase of silhouette score metrics is +32% for Scopus-long dataset, while the maximum increase of silhouette score metrics is +88% for Classic4 dataset, as shown in Fig. 6(a).

In terms of purity scores, the proposed models outperform WEClustering, as shown in Fig.5(b) and Table 2. WEClustering_K++ achieves a purity score of 0.975 in the Articles-253 dataset, slightly surpassing WEClustering_K's 0.971. However, in Classic4, WEClustering_K++ records a slightly lower score (0.89) than WEClustering_K's 0.911. Classic4-long shows a significant increase, with purity rising from 0.958 to 0.985. Scopus sees a reduction from 0.975 to 0.88, while Scopus-long experiences an increase from 0.722 to 0.76. The 20NG dataset demonstrates a notable improvement, with the purity score increasing from 0.534 to 0.701, while 20NG-long records a change from 0.397 to 0.321. WEClustering_A++ also shows consistent improvements, with Articles-253 increasing from 0.963 to 0.971, Classic4 experiencing an increase from 0.925 to 0.912, and Classic4-long showing a remarkable jump from 0.94 to 0.991. Scopus-long remains stable at 0.722, while the 20NG dataset sees a 6.1% rise. The 20NG-long dataset experiences the most substantial gain, improving from 0.291 to 0.46, reinforcing the efficacy of the proposed clustering technique. Also, an increase of median purity metrics of 0.4% and 0.8% is obtained for proposed WEClsutering_K++ and WEClustering_A++ compared to the state of art model, i.e. WEClustering_K[15] and WEClustering_A[15], respectively. In the case of WEClustering_K++, minimum change of purity score is -19.1% for 20NG-long dataset, while the maximum increase of purity score is +31.3% for 20NG dataset. In the case of WEClustering_A++, minimum change of purity score is -1.41% for Classic4 dataset, while the maximum increase of purity score is +58% for the 20NG-long, as shown in Fig. 6(b).

The WEClustering_K++ and WEClustering_A++ models also demonstrate significant advantages in terms of Adjusted Rand Index (ARI), as shown in Fig. 5(c) and Table 3. In the Articles-253 dataset, WEClustering_K++ attains an ARI of 0.985, a slight increase over WEClustering_K's 0.971. The Classic4 dataset records a rise from 0.932 to 0.96, whereas Classic4-long sees a slight drop from 0.96 to 0.915. The Scopus dataset showcases an improvement, with ARI increasing from 0.925 to 0.967, while Scopus-long advances from 0.71 to 0.76. The most prominent improvement is observed in the 20NG dataset, where the ARI score jumps from 0.434 to 0.601. Similarly, the 20NG-long dataset exhibits a sharp increase from 0.202 to 0.521. WEClustering_A++ also delivers strong performance gains, with Articles-253 experiencing a slight drop from 0.989 to 0.967. The Classic4 dataset shows only a minor decline, while Classic4-long sees an increase from 0.847 to 0.931. Scopus-long progresses from 0.672 to 0.8, while 20NG rises from 0.302 to 0.56. The highest improvement is observed in the 20NG-long dataset, where the ARI score jumps from 0.191 to 0.6, demonstrating the effectiveness of the proposed clustering method. Also, an increase of median ARI metrics of 7% and 11% is obtained for proposed WEClsutering_K++ and WEClustering_A++ compared to the state of art model, i.e. WEClustering_K[15] and WEClustering_A[15], respectively. In the case of WEClustering_K++, minimum change of ARI score is -4.69% for Classic4-long dataset, while the maximum increase of ARI score is +157.92% for 20NG-long dataset. In the case of WEClustering_A++, minimum change of ARI score is -2.23% for Articles-253 dataset, while the maximum increase of ARI score is +214.14% for the 20NG-long dataset, as shown in Fig. 6(c). Overall, the WEClustering_K++ and WEClustering_A++ models consistently deliver superior results compared to their baseline counterparts across all three-evaluation metrics.

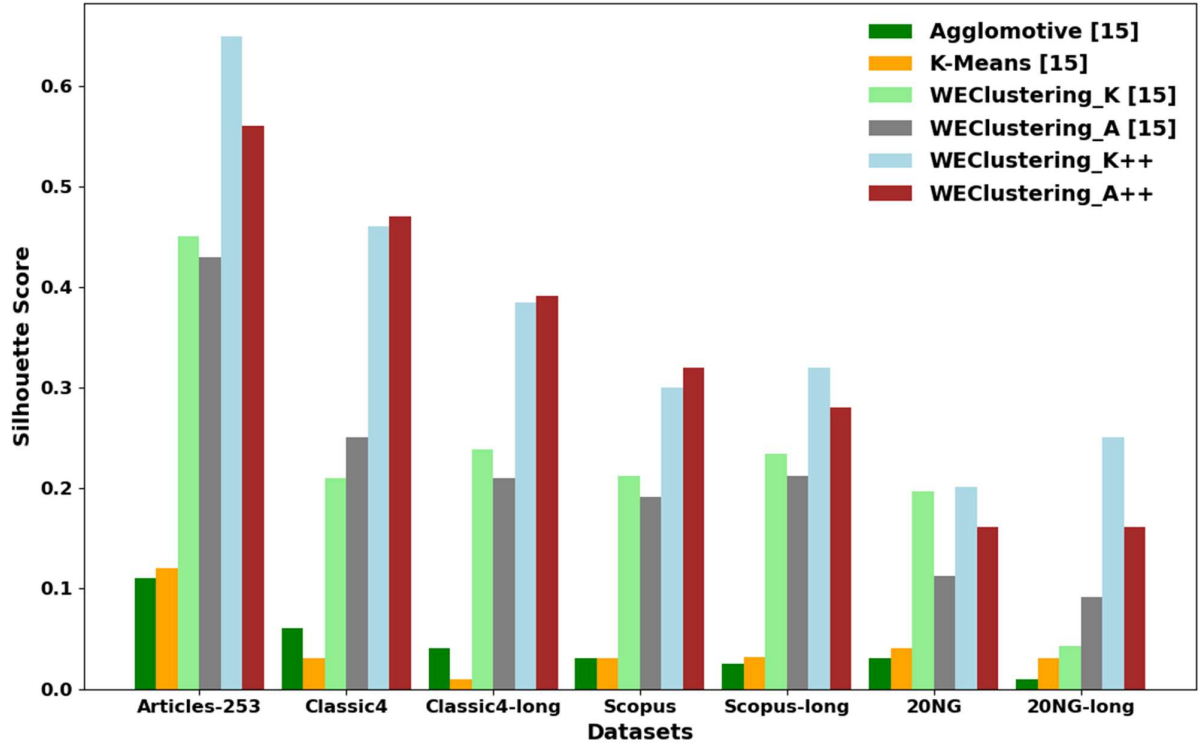

(a)

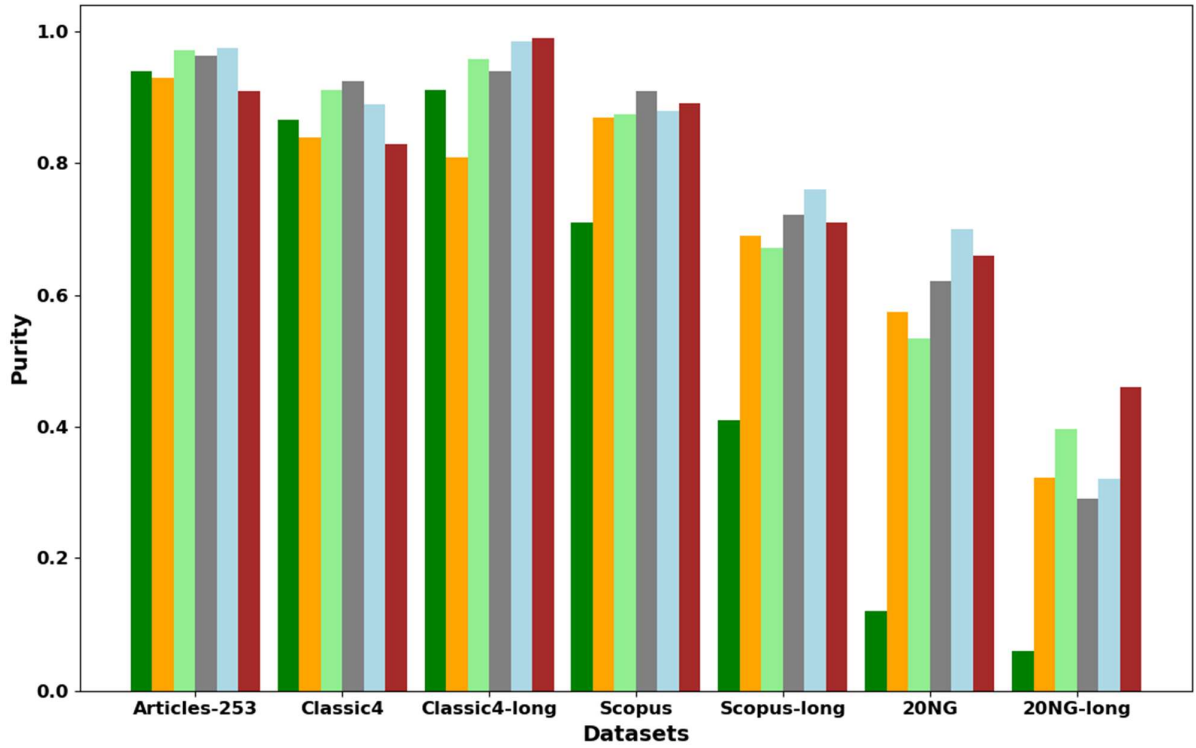

(b)

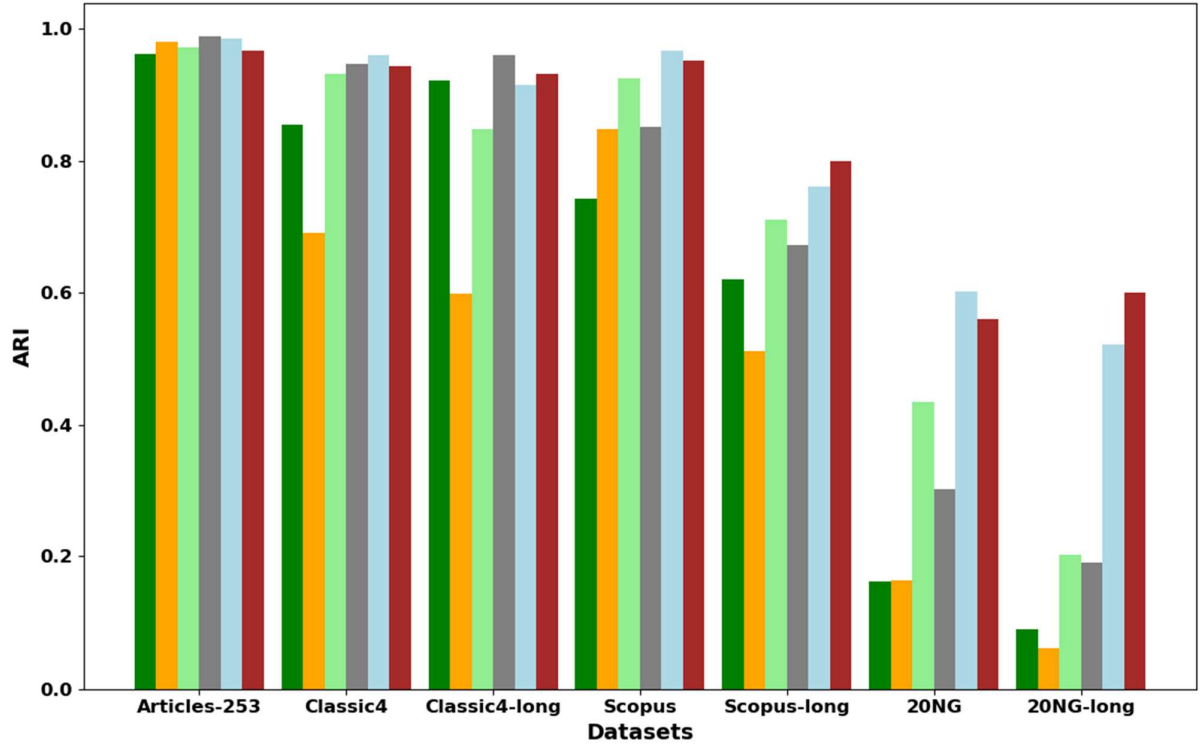

(c)

**Fig.5.** Variation of (a) Silhouette coefficient, (b) purity, and (c) ARI for different dataset and for different clustering techniques

Table 1. Silhouette coefficient of state-of-the-art models versus proposed model for different datasets

| Data | Agglomera-tive [15] | K_means [15] | WECluster-ing_K [15] | WECluster-ing_A [15] | WEClustering _K++ | WEClustering _A++ |
|---|---|---|---|---|---|---|
| **Articles-253** | 0.11 | 0.12 | 0.45 | 0.43 | 0.65 | 0.56 |
| **Classic4** | 0.06 | 0.03 | 0.21 | 0.25 | 0.46 | 0.47 |
| **Classic4-long** | 0.04 | 0.01 | 0.238 | 0.21 | 0.385 | 0.391 |
| **Scopus** | 0.03 | 0.03 | 0.212 | 0.191 | 0.3 | 0.32 |
| **Scopus-long** | 0.025 | 0.032 | 0.234 | 0.212 | 0.32 | 0.28 |
| **20NG** | 0.031 | 0.04 | 0.197 | 0.112 | 0.201 | 0.16 |
| **20NG-long** | 0.01 | 0.03 | 0.043 | 0.091 | 0.25 | 0.16 |

Table 2. Purity values of state-of-the-art models versus proposed model for different datasets

| Data | Agglomerative [15] | K_means [15] | WECluster-ing_K [15] | WECluster-ing_A [15] | WEClustering _K++ | WEClustering _A++ |
|---|---|---|---|---|---|---|
| Articles-253 | 0.94 | 0.93 | 0.971 | 0.963 | 0.975 | 0.971 |
| Classic4 | 0.866 | 0.84 | 0.911 | 0.925 | 0.89 | 0.912 |
| Classic4-long | 0.911 | 0.81 | 0.958 | 0.94 | 0.985 | 0.991 |
| Scopus | 0.71 | 0.87 | 0.975 | 0.91 | 0.88 | 0.892 |
| Scopus-long | 0.41 | 0.69 | 0.722 | 0.722 | 0.76 | 0.71 |
| 20NG | 0.12 | 0.574 | 0.534 | 0.622 | 0.701 | 0.66 |
| 20NG-long | 0.94 | 0.93 | 0.971 | 0.963 | 0.975 | 0.971 |

Table 3. ARI values of state-of-the-art models versus proposed model for different datasets

| Data | Agglomerative [15] | K_means [15] | WECluster-ing_K [15] | WECluster-ing_A [15] | WEClustering _K++ | WEClustering _A++ |
|---|---|---|---|---|---|---|
| Articles-253 | 0.961 | 0.98 | 0.971 | 0.989 | 0.985 | 0.967 |
| Classic4 | 0.854 | 0.69 | 0.932 | 0.947 | 0.96 | 0.943 |
| Classic4-long | 0.921 | 0.598 | 0.847 | 0.96 | 0.915 | 0.931 |
| Scopus | 0.743 | 0.847 | 0.925 | 0.851 | 0.967 | 0.952 |
| Scopus-long | 0.621 | 0.511 | 0.71 | 0.672 | 0.76 | 0.8 |
| 20NG | 0.162 | 0.164 | 0.434 | 0.302 | 0.601 | 0.56 |
| 20NG-long | 0.09 | 0.062 | 0.202 | 0.191 | 0.521 | 0.6 |

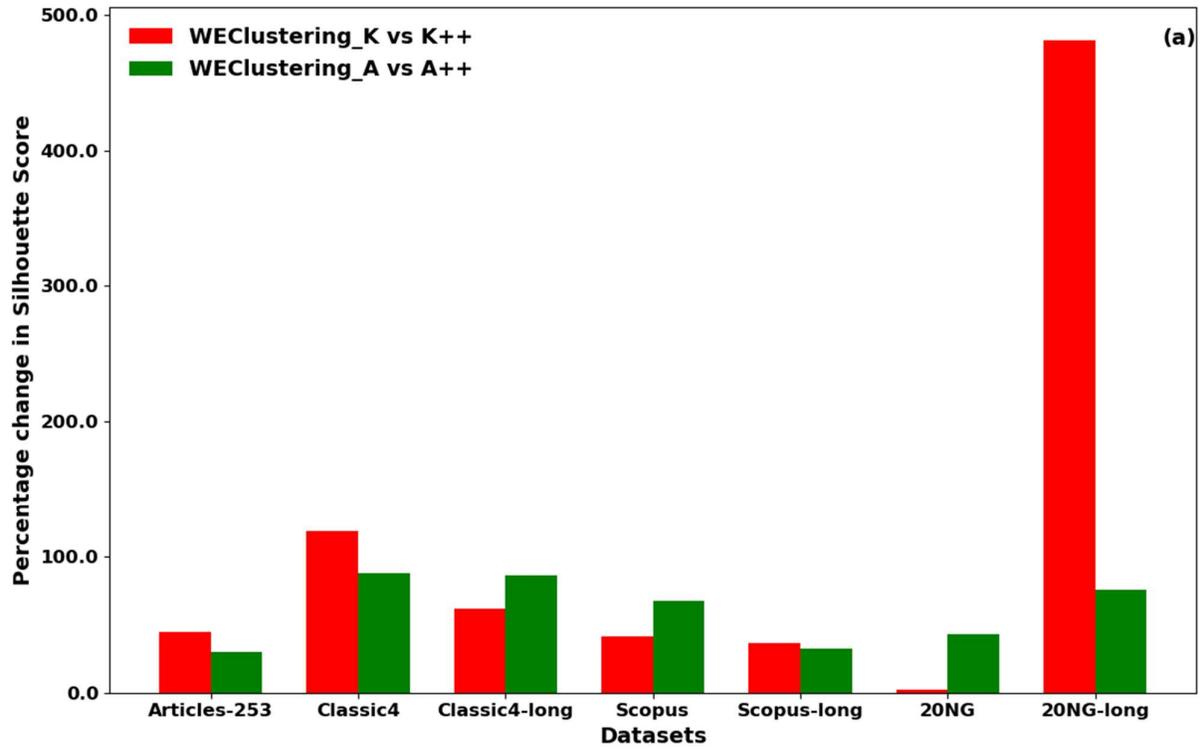
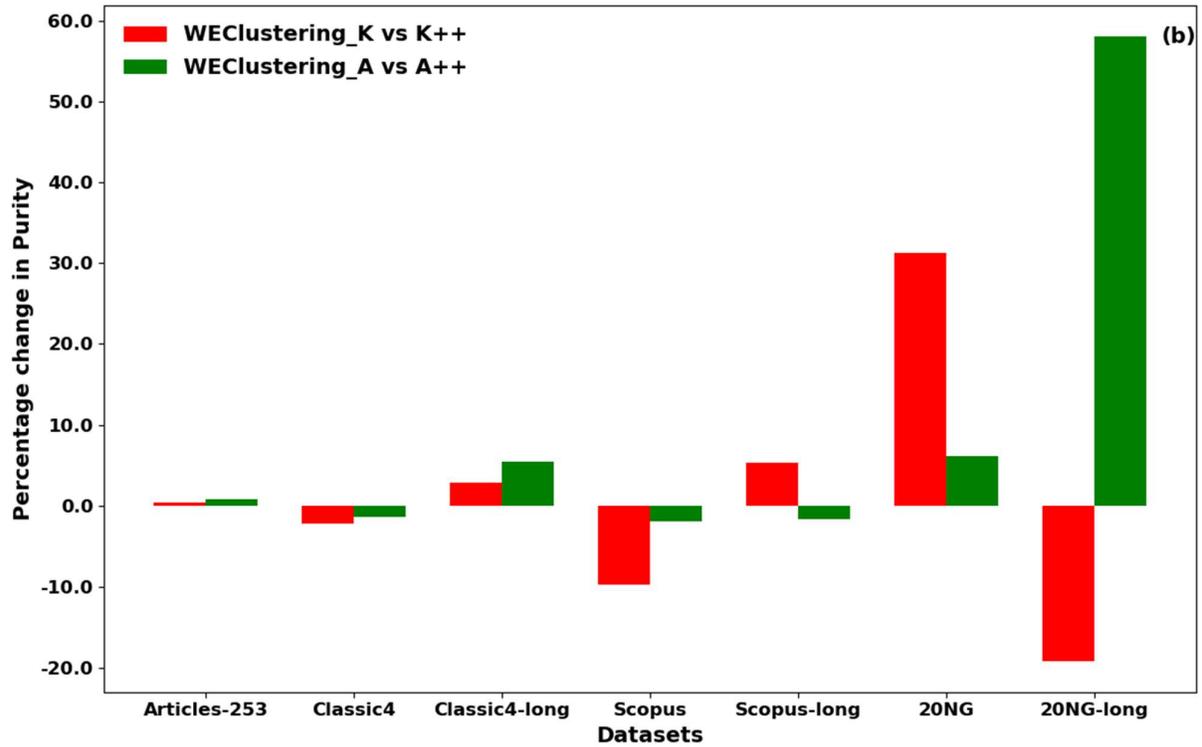

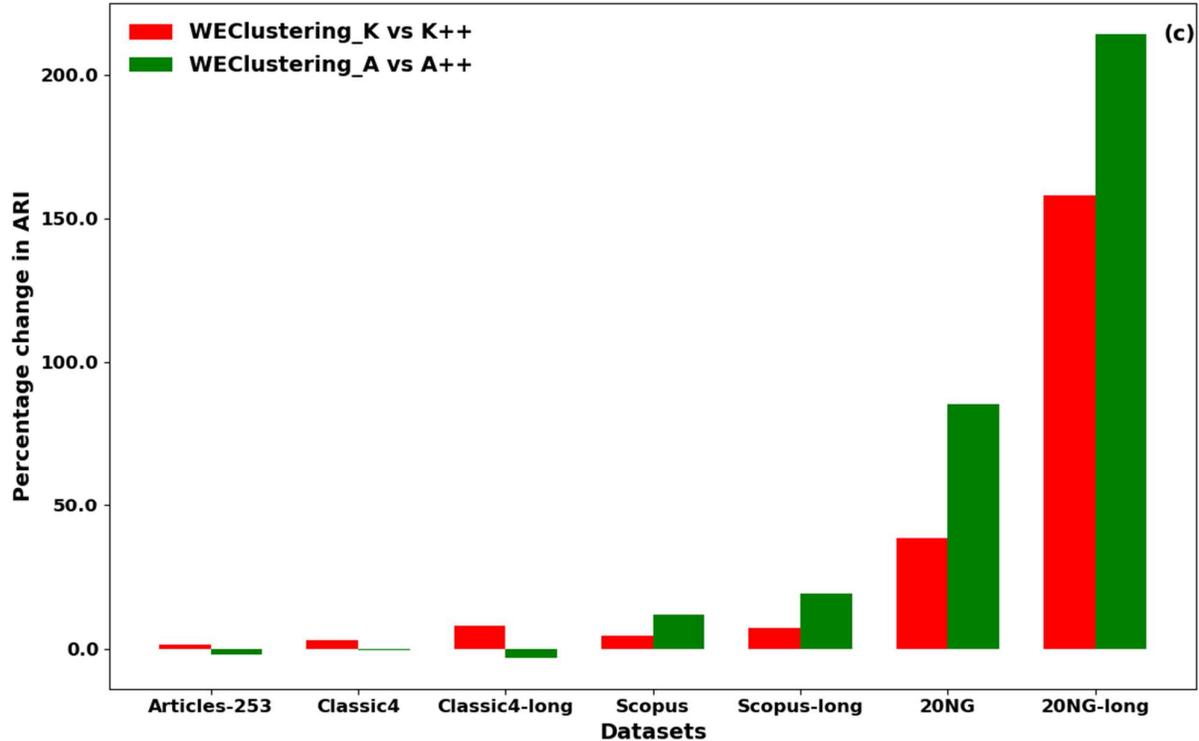

**Fig.6.** Percentage variations in (a) silhouette score, (b) purity, and (c) ARI of WEClustering_K [15] vs WEClustering_K++ and WEClustering_A [15] vs WEClustering_A++.

## Concluding remarks

The proposed WEClustering++ models significantly improve document clustering by integrating finetuned BERT embeddings, optimized dimensionality reduction, and enhanced clustering techniques. The model effectively captures semantic nuances by leveraging domain-specific contextual embeddings, leading to improved clustering quality compared to traditional and existing state-of-the-art methods. Experimental results across multiple benchmark datasets demonstrate that WEClustering++ achieves higher silhouette scores, purity, and adjusted Rand index (ARI), highlighting its ability to generate wellseparated and semantically meaningful clusters. The observed performance gains validate the importance of contextualized word embeddings and optimized clustering techniques in large-scale text mining tasks. The Proposed model may be further expanded to multilingual datasets, real-time clustering applications, and integration with adaptive learning models to further enhance scalability and efficiency.

## Acknowledgement

Authors would like to thank Shri. Y Dilip, Director, Aeronautical Development Establishment, Mr. Manjunath S M, Technology Director and Mr. Diptiman Biswas, Group Director for their support during the research work carried out at ADE, DRDO.